\def\year{2020}\relax
\documentclass[letterpaper]{article} 
\usepackage{aaai20}  
\usepackage{times}  
\usepackage{helvet} 
\usepackage{courier}  
\usepackage[hyphens]{url}  
\usepackage{graphicx} 
\usepackage{graphicx}  
\title{LGML: Logic Guided Machine Learning}

\author{ Joseph Scott,
Maysum Panju,
Vijay Ganesh\\
University of Waterloo\\
\{josesph.scott, mhpanju, vijay.ganesh\} @uwaterloo.ca
}

\begin{document}

\maketitle

\begin{abstract}
We introduce {\it Logic Guided Machine Learning} (LGML), a novel approach that symbiotically combines machine learning (ML) and logic solvers with the goal of learning mathematical functions from data. LGML consists of two phases, namely a {\it learning-phase} and a {\it logic-phase} with a corrective feedback loop, such that, the learning-phase learns symbolic expressions from input data, and the logic-phase cross verifies the consistency of the learned expression with known auxiliary truths. If inconsistent, the logic-phase feeds back "counterexamples" to the learning-phase. This process is repeated until the learned expression is consistent with auxiliary truth. Using LGML, we were able to learn expressions that correspond to the Pythagorean theorem and the sine function, with several orders of magnitude improvements in data efficiency compared to an approach based on an out-of-the-box multi-layered perceptron (MLP).
 
\end{abstract}

\section{Introduction to LGML}
\begin{figure}[t]
\begin{center}
    \centering
    \includegraphics[width=.95\linewidth]{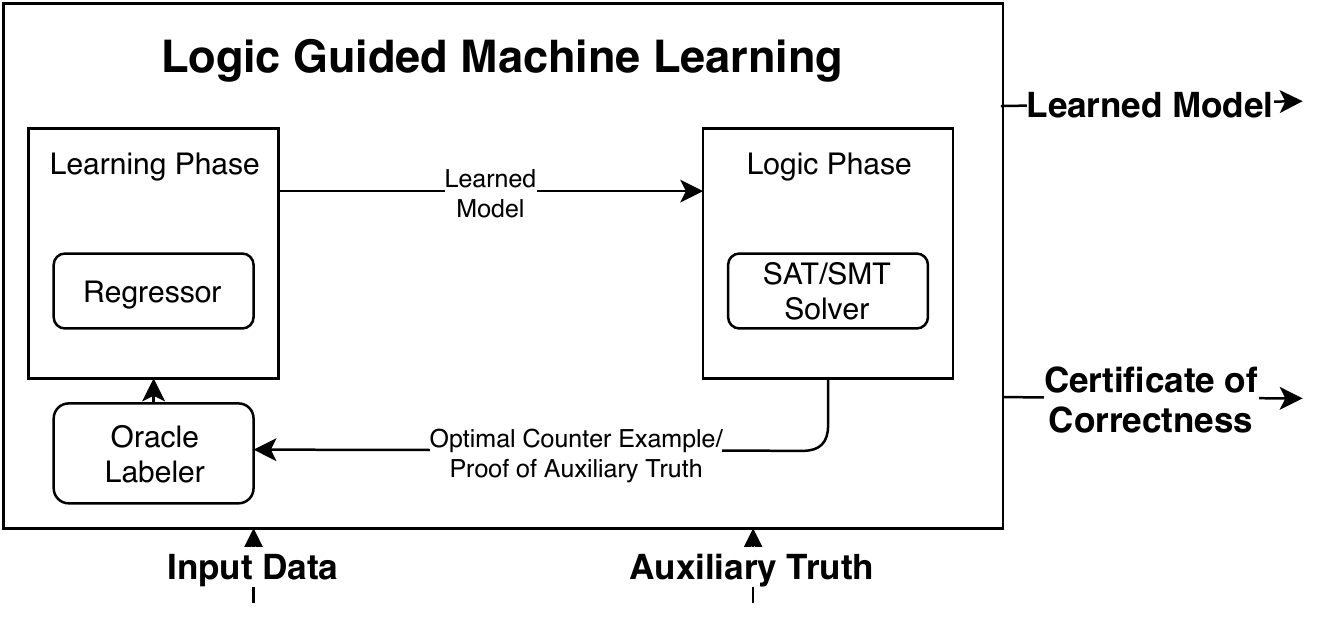}
    \caption{Architecture Diagram of of LGML}
  \label{fig:arch}
\end{center}
\end{figure}

We propose a new method of combining machine learning (the learning-phase) and logic solvers (the logic-phase) in a corrective feedback loop from the logic-phase to the learning-phase (see Figure \ref{fig:arch}), that we refer to as Logic Guided Machine Learning (LGML), aimed at learning mathematical expressions from data that are also consistent with previously known mathematical facts (auxiliary truths) expressed in a suitable fragment of mathematical logic. LGML takes the following as input: labeled data corresponding to an unknown target mathematical function $f$ to be learned and an auxiliary truth $\psi$ (theorems or invariants over $f$ in a suitable fragment of mathematics). LGML outputs a mathematical expression that fits the input data and is consistent with the input auxiliary truths.

The learning-phase of LGML fits a symbolic function $\hat{f}$, that approximates a target function $f$ (unknown to the system a priori), over the input data which is then fed to a logic solver (e.g., SAT/SMT solvers) along with the auxiliary truth $\psi$. If the symbolic expression $\hat{f}$ is inconsistent with the given auxiliary truth(s) $\psi$, the logic-phase feeds back the data point that violates the auxiliary truth the "most" (also referred to as the "strongest counterexample"). While the ground truth function $f$ underlying the dataset is unknown to the system, LGML does require access to an oracle to produce labels to the data point returned from the logic-phase. This process is repeated until $\hat{f}$ and $\psi$ are consistent with each other, i.e., $\hat{f} \models \psi$ (By this notation, we mean $\forall x. \psi(x,\hat{f}(x))$ is true).  

Those familiar with formal methods will immediately recognize a connection between LGML and the concept of formal verification. Given a program $P$ and a specification $\phi$ (written in a suitable fragment of mathematics), the goal of verification to check whether $P \wedge \neg \phi$ is satisfiable. If yes, then the satisfying assignment represents an input to the program $P$ that violates $\phi$. In LGML, the program $P$ is replaced by a symbolic representation $\hat{f}$ of the ML model and the (partial) specification is represented by auxiliary truth(s) $\psi$. If the input $\hat{f} \wedge \psi$ is deemed satisfiable by the logic-phase, the satisfying assignment thus produced represents the "strongest counterexample" which is fed back to the learning-phase to improve $\hat{f}$. One can also see a connection between LGML and the well-known verification method called Counterexample Guided Abstraction Refinement (CEGAR). In fact, LGML is inspired by~\cite{clarke2000counterexample}.

\begin{figure*}[t]

\begin{center}
    \centering
    \includegraphics[width=0.95\linewidth]{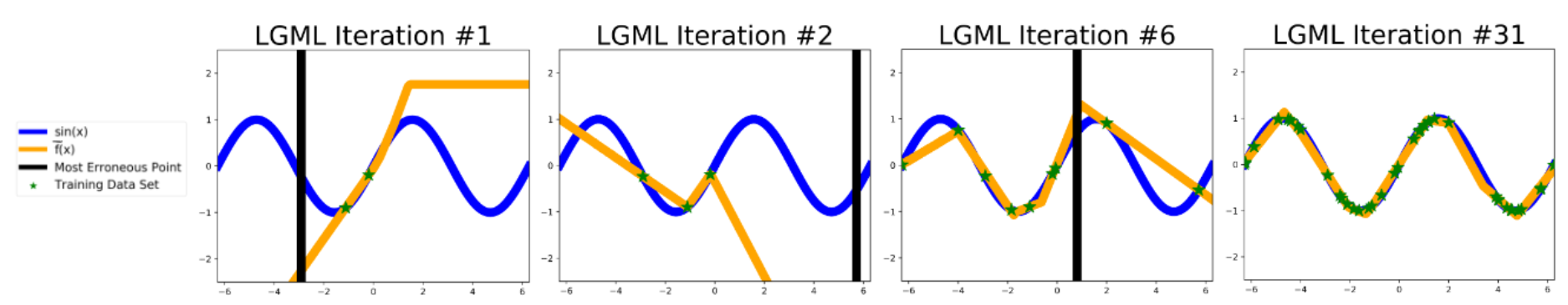}
    \label{fig:lgml:visual}
    \caption{Visualization of LGML for select iterations}

\end{center}

\end{figure*}

\begin{figure}[t]
\begin{center}
    \centering
    \includegraphics[width=.95\linewidth]{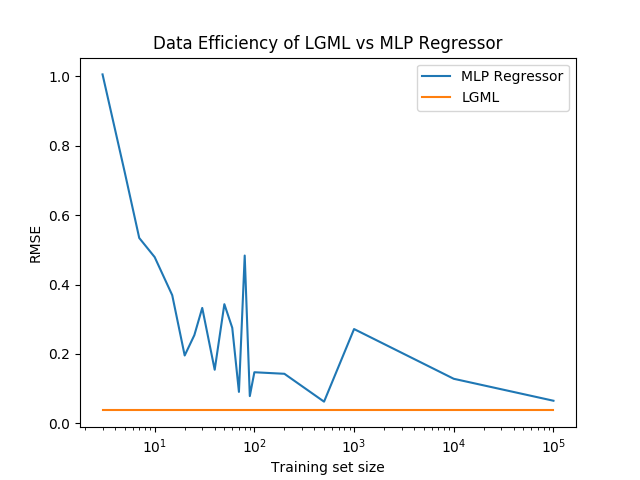}
    \caption{RMSE of LGML vs. standard MLP regressor.}
    \label{fig:lgml:err}
\end{center}
\end{figure}

\section{Input/Output Interface of LGML}
 
The input to LGML is a labeled dataset that relates the inputs and output of some function $f$ whose symbolic representation is not known to the system, and an auxiliary theorem or truth $\psi$ known to be true for the inputs and output $f$. For example, consider $f(a,b)$ to be the function that takes as input the length $a$ and $b$ of the sides of a right-angled triangle and computes the length of its hypotenuse using the Pythagorean theorem. Then, one possible $\psi$ is the triangle inequality $a+b > f(a,b)$. Even if one may not quite know the statement of the Pythagorean theorem, one may have a large sample of right-angled triangles (input data set) and know the statement of the triangle inequality (auxiliary truth). Combinatorial mathematicians and physicists routinely have access to such data sets and auxiliary truths and want to learn some previously unknown function or invariant over such data \cite{larson2017automated}. 

{\bf Output of LGML:} LGML outputs some function $\hat{f}$ that fits the labeled dataset and is consistent with $\psi$. Unlike classical regressors, LGML starts with a small dataset and augments it with additional points during runtime, obtained via a logical-phase. 

\section{Phases of LGML}

{\bf The learning-phase of LGML:} 
The learning-phase applies a regression algorithm on the current dataset to obtain a math expression $\hat{f}$ that fits the input data. Examples of regression algorithms include symbolic regression methods and deep neural networks. Importantly, we require the regressor to fit all data points with near-perfect accuracy. 



{\bf The logic-phase of LGML:} 
The logic-phase checks whether the learned function $\hat{f}$ is consistent with the input auxiliary truth $\psi$, denoted as $\hat{f} \models \psi$. As long as this is not the case, LGML computes the 'strongest' counterexample resulting in inconsistency, inserts it into the training set, and repeats the learning-phase on the augmented training data. First note that $\psi$ is a logical equation in terms of the feature space and the unknown function $f$, or more precisely, in terms of the output of $\hat{f}$ that approximates $f$. Hence, in its logic-phase, LGML constructs a satisfiability query that checks whether $\hat{f} \models \psi$. If inconsistent, a counterexample is extracted from the logic solver. LGML finds the 'strongest' such counterexample and returns it to the learning-phase.

{\bf Auxiliary Truth as a Satisfiability Query:}

We weaken the auxiliary truth $\psi$ (a quantifier-free mathematical formula) to be written with an error term $\epsilon$, denoted as $\psi_\epsilon$. Suppose the auxiliary truth $\psi$ is of the form $\alpha = \beta$, where $\alpha$ and $\beta$ are well-formed symbolic expressions. Then we construct the query: \[ \lnot \psi_\epsilon := |\alpha[f/\hat{f}] - \beta[f/\hat{f}]| > \epsilon\] Further, if $\psi$ is of the form of an inequality, (e.g $\alpha > \beta$), then:  \[ \lnot \psi_\epsilon :=  \beta[f/\hat{f}] - \alpha[f/\hat{f}] > \epsilon \] where the $/$ is the logical substitution. The satisfiability is checked via a floating-point (FP) SMT solver for a fixed $\epsilon$. We model $\hat{f}$ with 16-bit precision and use CVC4 as the backend  FP SMT solver.

\vspace{0.10cm}
{\bf Extracting Proof and the Strongest Erroneous Point:} We use a variant of the well-known bisection method to find both proofs and the strongest erroneous points with respect to an auxiliary truth. 

By making multiple queries to an FP SMT solver, we compute $\epsilon^*$ such that 

\vspace{0.10cm}
\noindent{for all}  $\epsilon > \epsilon^*$ \[\hat{f} \models \psi_{\epsilon}\] and for all $\epsilon \leq \epsilon^*$, \[\hat{f} \not \models \psi_{\epsilon}\]

For a target machine error $\rho$, we first query $\epsilon=\rho$. The terminating condition of LGML is the proof of $\hat{f} \models \psi_\rho$. We exponentially increase $\epsilon$ by doubling its value until a SAT result (i.e $\hat{f} \not\models \psi_\epsilon$). An UNSAT result (i.e $\hat{f} \models \psi_\epsilon$) establishes an interval containing $\epsilon^*$, which we narrow using the bisection method until convergence on $\epsilon^*$ and as consequently the strongest erroneous point.

\section{Evaluation}

We empirically tested LGML on two tasks: learning the function $f$ corresponding to Pythagorean theorem and the sine function. For brevity, we focus on sine function here. The auxiliary truth we use $\psi:=\sin^2(x) + \cos^2(x) = 1$, and  the satisfiability query that we use is $|\hat{f}(x)^2 + \left(\frac{d}{dx} \hat{f}(x)\right)^2 -1| > \epsilon$, for various $\epsilon$ as described until $\epsilon^*$ is found. As the base learning model, we use an MLP regressor with two hidden layers of 3 nodes each. Figure \ref{fig:lgml:visual} visualizes select iterations of LGML. As can be seen, the LGML learns the Sine function almost perfectly. 

We evaluate LGML at the 30th iteration (for a total of 32 training points) with a testing set of 10,000 points and compute an RMSE of 0.037. As a baseline, an MLP regressor was trained to learn the same function without using LGML, but given increasingly large training sets. Figure \ref{fig:lgml:err} demonstrates that the LGML model achieved lower error scores on learning $f(x)=\sin(x)$ using just 32 training points than the non-LGML MLP model when given even 100,000 points.

\bibliographystyle{aaai}
\bibliography{ref}

\end{document}